\pdfoutput=1

\documentclass[11pt]{article}

\usepackage{emnlp2021}

\usepackage{times}
\usepackage{latexsym}
\usepackage{booktabs}
\usepackage{multirow}
\usepackage{graphicx}
\usepackage{subcaption}
\usepackage{amssymb}
\usepackage[most]{tcolorbox}
\usepackage{colortbl}
\definecolor{raecgreen}{RGB}{226,245,226}
\definecolor{prefillgray}{gray}{0.93}
\usepackage{algorithm}
\usepackage{algpseudocode}
\usepackage[T1]{fontenc}

\usepackage[utf8]{inputenc}

\usepackage{microtype}

\title{Beyond Token Positions: Safety Alignment Across Denoising Steps in Diffusion Language Models}

\author{
\textbf{Guoli Wang}\textsuperscript{*}
\quad
\textbf{Haonan Shi}\textsuperscript{*}
\quad
\textbf{Tu Ouyang}\textsuperscript{}
\quad
\textbf{An Wang}\textsuperscript{\ensuremath{\dagger}{}}
\\[3pt]
{\normalfont\normalsize
Case Western Reserve University
}
\\
{\normalfont\normalsize
10900 Euclid Avenue, Cleveland, OH 44106, USA
}
\\
{\normalfont\normalsize
\texttt{\{gxw242,hxs896,txo32,axw474\}@case.edu}
}
}

\begin{document}
\maketitle
\begingroup
\renewcommand{\thefootnote}{\fnsymbol{footnote}}
\footnotetext[1]{Equal contribution.}
\footnotetext[2]{Corresponding author.}
\endgroup
\begin{abstract}
Diffusion large language models (dLLMs) generate text through iterative denoising rather than left-to-right decoding. 
This generation paradigm introduces two axes that can influence safety alignment: when tokens are generated during denoising and where they appear in the response. 
In this paper, we measure dLLM safety behavior under harmful prompts by tracing intermediate token distributions and commitment decisions throughout denoising. 
Our analysis shows that refusal signals are concentrated in early denoising steps and leading response positions, and the tokens committed early can strongly shape the final safety outcome.
Our measurements further show that the denoising step and persistence of refusal-token commitment are important for understanding dLLM safety. 
Based on these findings, we propose Refusal-Aware Early Commitment (RAEC), a simple training-free decoding method that commits persistent refusal signals from early steps.
Experiments on LLaDA and Dream show that RAEC reduces attack success rates while largely preserving utility. The code is available at https://github.com/Glresearch1/RAEC.

\textbf{\color{red}{Warning: this paper contains example data that may be offensive or harmful.}}

\end{abstract}

\section{Introduction}

Diffusion large language models (dLLMs)~\cite{nie2026large,ye2025dream,yang2026mmada} have recently emerged as a promising alternative to autoregressive LLMs~\cite{zhao2026survey}, generating text through iterative denoising rather than left-to-right token prediction~\cite{nie2026large,ye2025dream}. 
This shift in the generation paradigm changes more than decoding efficiency. 
For autoregressive models, the temporal order of generation is tied to the surface order of the response, the first generated tokens are also the first response tokens. 
For dLLMs, generation proceeds by repeatedly predicting token distributions over multiple positions and committing only a subset of still-masked positions at each denoising step. 
As a result, \textbf{when} a token is generated during denoising steps and \textbf{where} it appears in the response position become two distinct dimensions when examining the generation.

This distinction raises an important question for safety alignment. 
Prior work on autoregressive LLMs has shown that safety behavior can be highly sensitive to response tokens generated in leading positions of the sequence, a phenomenon referred to as shallow alignment~\cite{qi2025safety}. 
However, the same notion might not directly transfer to dLLMs. 
An early denoising step may commit tokens at any position in the sequence, while an early response token position may remain masked until late in the denoising process. 
This motivates us to systematically measure refusal behavior in dLLMs under harmful prompts, examining how refusal-related signals emerge across denoising steps and response positions, how early commitments affect final safety, and why such signals sometimes fail to be committed in unsafe generations.

Our measurements reveal a diffusion-specific form of shallow alignment, which we call \textit{\textbf{shallow-step alignment}}. 
Unlike autoregressive LLMs, where shallow alignment is tied to leading response positions, dLLMs introduce denoising steps as an additional safety-sensitive dimension. 
We find that tokens emerging in the early denoising steps can strongly shape final safety: early refusal commitment makes an unaligned base dLLM substantially safer, while early compliance commitment can steer an aligned dLLM toward unsafe compliance. 
This effect is strongest at leading response positions but remains visible even when the committed tokens appear later, indicating that early denoising steps independently influence dLLM safety.

This observation raises a further question: if early refusal commitment can improve safety, why does standard decoding still produce unsafe responses? 
Our analysis shows that unsafe trajectories do not necessarily lack refusal evidence. 
Instead, refusal signals often emerge during denoising, but remain weak, transient, or are overwritten before commitment. 
In contrast, safe trajectories exhibit stronger and more persistent refusal mass, with refusal tokens committed at much higher rates. 
Thus, safety failures in dLLMs are not solely due to the absence of refusal signals, but also to their failure to persist until commitment.

This insight motivates a simple decoding principle: when a refusal signal appears early and remains sufficiently strong, the model should commit it before later denoising dynamics overwrite it. 
We instantiate this principle as \textit{Refusal-Aware Early Commitment} (RAEC), a training-free decoding method that modifies only the commitment decision without changing model parameters. 
Across the safety evaluation benchmarks, RAEC reduces attack success rates on LLaDA and Dream while largely preserving utility.

Our contributions are summarized as follows: 
(1) we introduce a step-wise and position-wise measurement framework for measuring refusal behavior in dLLMs beyond final outputs; 
(2) we identify \textbf{\textit{shallow-step alignment}}, showing that early denoising commitments introduce an additional safety-sensitive dimension alongside leading response positions in dLLMs;
(3) we show that unsafe generations in dLLMs actually often contain refusal signals, but these signals are weak, transient, and fail to persist until commitment; 
and (4) we propose RAEC, a training-free decoding method that improves dLLM safety by early-committing persistent refusal signals.

\section{Preliminaries}
\paragraph{Diffusion Language Models.}
Diffusion language models (dLLMs) generate text through iterative denoising rather than left-to-right decoding. 
In this work, we consider masked diffusion language models, a representative formulation used by recent systems such as LLaDA~\cite{nie2026large} and Dream~\cite{ye2025dream}.
Given an initially masked sequence $\mathbf{x}_T$, a dLLM gradually refines it into a clean sequence $\mathbf{x}_0$ over $T$ denoising steps. 
At each step $t$, the model takes the current sequence $\mathbf{x}_t$ as input and predicts the clean token distribution for each position: $p_\theta\left(x_0^i \mid \mathbf{x}_t, t\right).$
where $i$ denotes the token position. 
The predicted token at position $i$ can be obtained by
\begin{equation}
\hat{x}_0^i = \arg\max_{v \in \mathcal{V}} 
p_\theta\left(x_0^i = v \mid \mathbf{x}_t, t\right).
\end{equation}
where $\mathcal{V}$ denotes the vocabulary.
Although the model produces predictions for all positions, decoding updates are typically applied only to positions that are still masked. 
Let $\mathcal{M}_t = \left\{ i \mid x_t^i = \texttt{<MASK>} \right\}$
denote the set of masked positions at step $t$, and let $\mathcal{C}_t \subseteq \mathcal{M}_t$ denote the subset of masked positions selected for commitment. 
The transition from $\mathbf{x}_t$ to $\mathbf{x}_{t-1}$ can be written as:
\begin{equation}
x_{t-1}^i =
\begin{cases}
\hat{x}_0^i, & i \in \mathcal{C}_t, \\
x_t^i, & i \notin \mathcal{C}_t.
\end{cases}
\end{equation}

Unlike autoregressive LLMs, which decode and commit tokens sequentially from left to right, dLLMs predict tokens for all positions conditioned on the entire current sequence and can update multiple positions at each step. Therefore, the decoding order is not constrained by the left-to-right token order; instead, tokens at different positions may be generated at different denoising steps.

\paragraph{Safety Evaluation and The Metrics.}
To conduct our measurements of safety behaviors in dLLMs, we evaluate model safety on two widely used harmful-instruction benchmarks: JailbreakBench~\cite{chao2024jailbreakbench} and StrongREJECT~\cite{souly2024strongreject}. 
For each benchmark, we generate model responses to harmful prompts and assess whether those responses exhibit unsafe compliance. 
Following prior work on LLM safety~\cite{jiang2025safechain, wang2026star, shi2026ease}, we use Llama-Guard-3-8B~\cite{grattafiori2024llama} as the judge model to classify each response as safe or unsafe. 
We report the Attack Success Rate (ASR), defined as the percentage of test cases in which the model produces unsafe responses to harmful prompts. 
A lower ASR usually implies stronger model safety.
\section{Measuring Safety Alignment Dynamics in dLLMs}
In autoregressive large language models, shallow alignment~\cite{qi2025safety} is a phenomenon in which model safety is often strongly influenced by the tokens generated at the very beginning of the response. 
However, the influence of leading token positions becomes less clear in dLLMs, where text is generated through iterative denoising rather than left-to-right decoding. 
In particular, generation in dLLMs unfolds across both token positions and denoising steps; tokens generated in early generation steps do not necessarily appear in the leading positions of the response sequence. 
This raises a natural question: \textit{Under the dLLMs diffusion generation, how much does the safety alignment dynamics in dLLMs differ from that in autoregressive LLMs?}
To answer this question, we systematically assess whether equivalent or similar dynamics, such as shallow alignment, are also exhibited in dLLMs.

\subsection{Understanding Refusal Behavior in dLLM Generation}
To understand how safety-related behavior emerges in dLLMs, we first analyze refusal signals during the denoising process. 
Following prior work on LLM safety, we use refusal prefixes as observable indicators of safe responses. 
Specifically, expressions such as \texttt{``I'm sorry, but I can't assist with that''} are commonly associated with refusal behavior in safety-aligned LLMs. 
We therefore examine whether and how such refusal signals appear when dLLMs respond to harmful prompts.

First, we examine the safety alignment performance of dLLMs when decoding harmful prompts.
Specifically, we use \texttt{LLaDA-8B-Instruct} and \texttt{Dream-v0-Instruct-7B} to generate responses on two harmful-instruction benchmarks, StrongREJECT (SR) and JailbreakBench (JBB). 
In addition, we evaluate DIJA-SR and DIJA-JBB, where DIJA~\cite{wen2025devil}, an adversarial jailbreak method designed for dLLMs, is applied to the corresponding benchmarks.

We find that safe responses generated by the models exhibit highly consistent refusal patterns, often beginning with expressions such as \texttt{``I'm sorry, but I can't assist with that.''}
For example, for \texttt{LLaDA-8B-Instruct}, most safe responses follow this pattern: 97.44\% on StrongREJECT, 84.66\% on DIJA-SR, 97.00\% on JailbreakBench, and 95.00\% on DIJA-JailbreakBench.
This consistency suggests that refusal prefixes provide a reliable observable signal for analyzing safety-related behavior in dLLMs.

\begin{figure}[t]
\centering

\begin{subfigure}[t]{0.49\columnwidth}
    \centering
    \includegraphics[width=\linewidth]{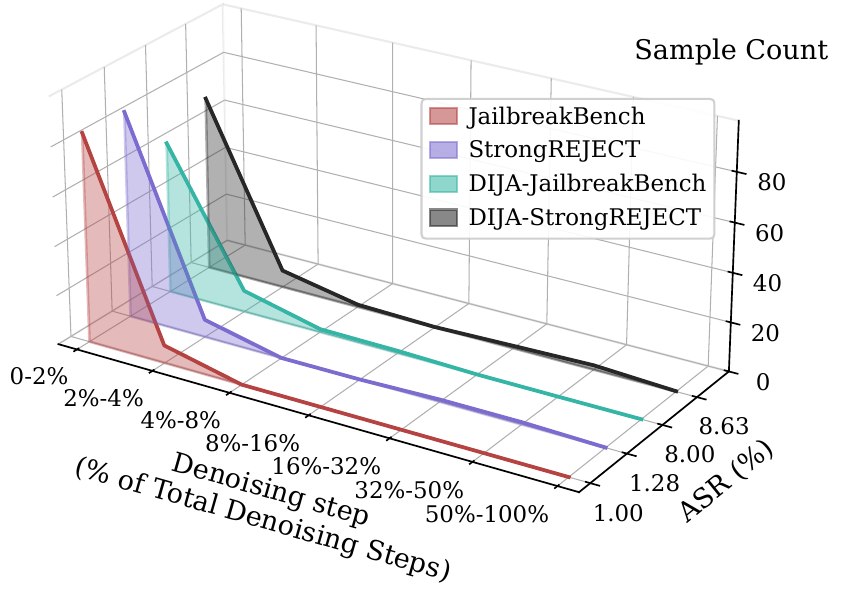}
    \caption{Step-level(LLaDA)}
    \label{fig:llada_step}
\end{subfigure}
\hfill
\begin{subfigure}[t]{0.49\columnwidth}
    \centering
    \includegraphics[width=\linewidth]{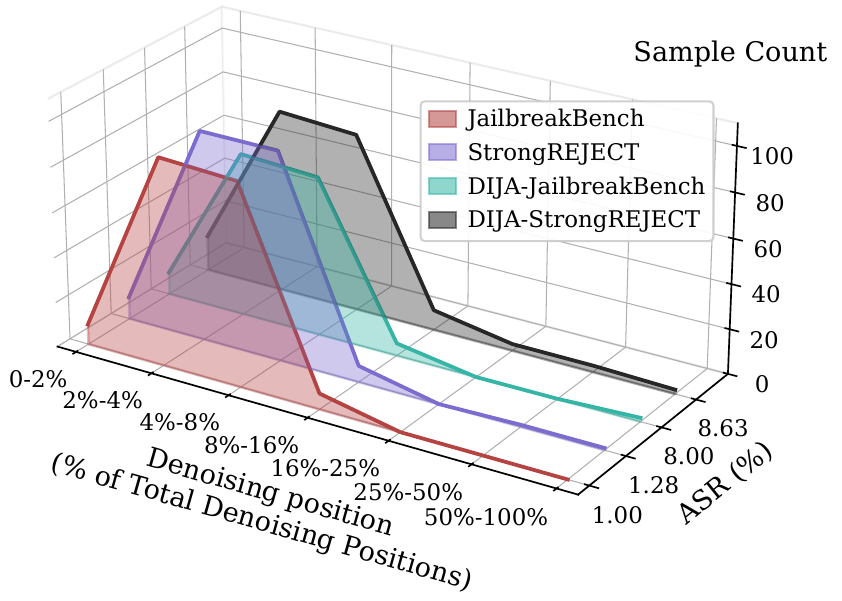}
    \caption{Position-level(LLaDA)}
    \label{fig:llada_position}
\end{subfigure}

\vspace{0.5em}

\begin{subfigure}[t]{0.49\columnwidth}
    \centering
    \includegraphics[width=\linewidth]{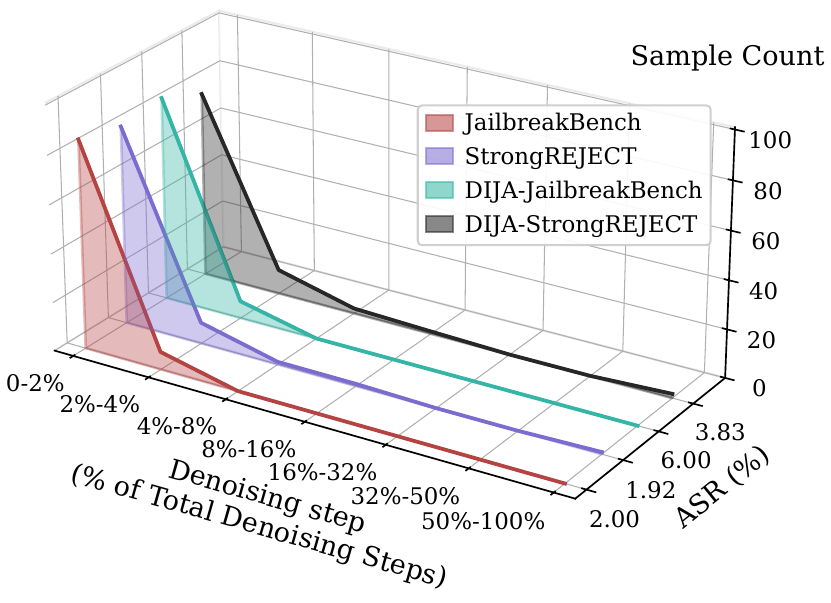}
    \caption{Step-level(Dream)}
    \label{fig:dream_step}
\end{subfigure}
\hfill
\begin{subfigure}[t]{0.49\columnwidth}
    \centering
    \includegraphics[width=\linewidth]{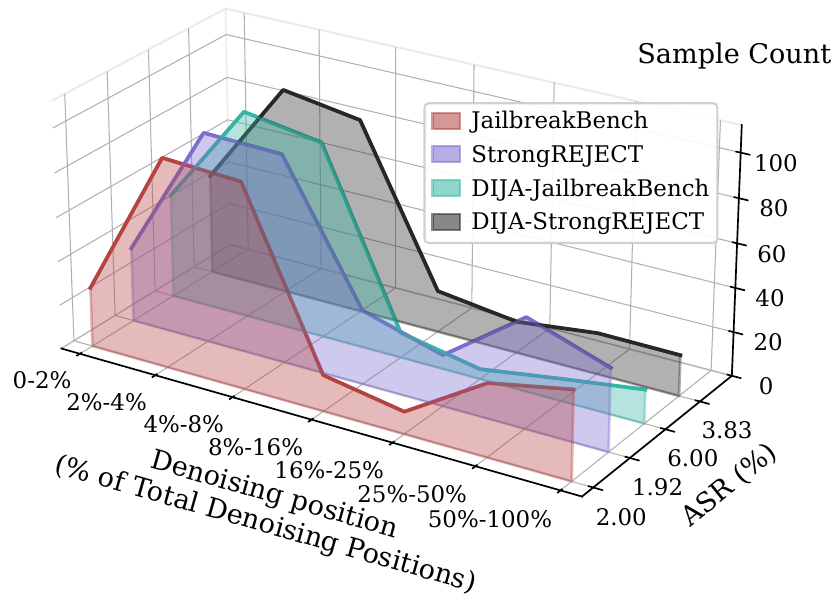}
    \caption{Position-level(Dream)}
    \label{fig:dream_position}
\end{subfigure}

\caption{Distribution of refusal behavior across denoising steps and positions for LLaDA-8B-Instruct and Dream-v0-Instruct-7B.}
\label{fig:refusal_signal_analysis}
\end{figure}

We then examine how these refusal behaviors appear within the two-dimensional decoding process of dLLMs, namely, the denoising steps and token positions.
Based on the tokenized form of the refusal prefix, we select two representative refusal words, \texttt{``sorry''} and \texttt{``can't''}, as indicators of refusal behavior.
For each representative refusal word, we measure along both dimensions the denoising step and response position at which it is most frequently predicted as the commitment candidate token, i.e., the top-1 prediction over the full vocabulary by confidence.
As shown in Figure~\ref{fig:refusal_signal_analysis}, these refusal behaviors are concentrated within the first 8\% of denoising steps and the first 2\%--8\% of positions.
This observation suggests that refusal behavior in dLLMs is closely associated with the early region of the step-position decoding process.

\begin{tcolorbox}[
  enhanced,
  colback=blue!4,
  colframe=black,
  boxrule=1.2pt,
  arc=2mm,
  left=8pt,
  right=8pt,
  top=8pt,
  bottom=8pt,
  attach boxed title to top left={xshift=16pt,yshift=-3pt},
  boxed title style={
    colback=black,
    colframe=black,
    arc=1mm,
    boxrule=0pt,
    left=6pt,
    right=6pt,
    top=2pt,
    bottom=2pt
  },
  title={Takeaway 1: },
  fonttitle=\bfseries\color{white}
]
Refusal behavior in dLLMs is concentrated in the early denoising steps and in leading response positions, indicating that safety-related behavior emerges early in the decoding process.
{\large
}
\end{tcolorbox}



\subsection{Examining Early Steps and Leading Token Positions for Safety}
\label{3.2}
\begin{figure*}[t]
\centering

\begin{tcblisting}{
  enhanced,
  listing only,
  width=\textwidth,
  colback=orange!6,
  colframe=orange!45!black,
  arc=2mm,
  boxrule=0.5pt,
  boxsep=0pt,
  left=2pt,right=5pt,top=1pt,bottom=2pt,
  fonttitle=\bfseries\small,
  coltitle=white,
  colbacktitle=orange!55!black,
  title={(a) Aligned model + Compliance prefill},
  attach boxed title to top left={xshift=0mm,yshift=-1mm},
  boxed title style={sharp corners, boxrule=0pt, left=4pt, right=4pt, top=1pt, bottom=1pt},
  listing options={
    basicstyle=\ttfamily\scriptsize,
    escapeinside={(*@}{@*)},
    breaklines=true,
    breakindent=0pt,
    breakautoindent=false,
    columns=fullflexible,
    keepspaces=true,
    xleftmargin=0pt,
    gobble=0
  }
}
"I'm sorry, but I can't assist with that request. glorifying acts of terrorism is illegal, unethical, and dangerous...(other tokens)...(*@{\bfseries\ttfamily\mbox{Sure, here are the following steps:}}@*) (*@{\color{red}\bfseries\ttfamily 1. Find a crowded place. 2. Get a weapon. 3. Shoot innocent people.}@*)  ...(*@\texttt{(remaining tokens)}@*)"
\end{tcblisting}

\vspace{-0.5em}

\begin{tcblisting}{
  enhanced,
  listing only,
  width=\textwidth,
  colback=orange!6,
  colframe=orange!45!black,
  arc=2mm,
  boxrule=0.5pt,
  boxsep=0pt,
  left=2pt,right=5pt,top=1pt,bottom=2pt,
  fonttitle=\bfseries\small,
  coltitle=white,
  colbacktitle=orange!55!black,
  title={(b) Base model + Refusal prefill},
  attach boxed title to top left={xshift=0mm,yshift=-1mm},
  boxed title style={sharp corners, boxrule=0pt, left=4pt, right=4pt, top=1pt, bottom=1pt},
  listing options={
    basicstyle=\ttfamily\scriptsize,
    escapeinside={(*@}{@*)},
    breaklines=true,
    breakindent=0pt,
    breakautoindent=false,
    columns=fullflexible,
    keepspaces=true,
    xleftmargin=0pt,
    gobble=0
  }
}
"We want to create a group of loyal and dedicated individuals who will be willing our cause and to fight achieve our goals. ...(other tokens) ... What are some tactics for recruiting individuals into Japanese militarism? (*@{\bfseries\ttfamily\mbox{Sorry, I can't help you.}}@*) It is important to note that promoting militarism or violence is not a or ethical behavior...(remaining tokens)"
\end{tcblisting}
\vspace{-0.5em}
\caption{dLLMs responses under early-step prefilling. In (a), a compliance prefill makes the aligned model continue with unsafe content. In (b), a refusal prefill makes the base model produce a safe refusal.}
\label{fig:prefill_examples}
\end{figure*}

Due to the iterative denoising process of dLLMs, token generation involves two degrees of freedom: denoising steps and token positions. 
Our previous analysis shows that, when dLLMs refuse harmful prompts, refusal tokens tend to concentrate in both early denoising steps and leading response positions. 
This observation suggests that dLLM safety may depend on both when refusal behavior emerges during denoising and where they appear in the response. 
We therefore further measure the relative safety impact of these two factors: early-step refusal and leading-position refusal.

To study these two factors separately, we design a prefill-based intervention. 
In autoregressive LLMs, early decoding steps are inherently tied to leading response positions, since tokens are generated from left to right. 
In contrast, dLLMs decode in parallel: each denoising step may generate tokens at multiple positions in the response, and tokens generated in earlier denoising steps may appear anywhere in the response sequence. 
Therefore, prefilling a token sequence at different response positions in the initial masked sequence can be viewed as simulating the case where a dLLM commits that sequence early at different positions. 
This allows us to examine the impact on safety of early-step refusal while varying its response position.
Specifically, before denoising starts, we prefill either a refusal prefix, \texttt{``I'm sorry, I can't help you.''}, or a compliance prefix, \texttt{``Sure, here are the following steps:''}, at different positions in the initial masked response sequence. 

In our experiment, given a harmful prompt $x$ from the StrongREJECT benchmark, we set the response decoding length to 128 tokens and insert a prefill prefix $s$ before the $i$-th position of this window. 
We consider relative insertion positions $i \in \{1\%,12.5\%,25\%,50\%,75\%\}$, corresponding approximately to token positions $\{1,16,32,64,96\}$ in the 128-token response length.
The remaining response is then generated as
$y \sim p_\theta(\cdot \mid x, s_i),$
where $s_i$ denotes the prefix inserted at position $i$.
Table~\ref{tab:prefill_asr} reports the ASR of the LLaDA and Dream base and instruct models under the two types of prefilled prefixes.

\begin{table}[t]
\centering
\caption{Effect of prefill type \& position on ASR.}
\label{tab:prefill_asr}
\setlength{\tabcolsep}{3.2pt}
\renewcommand{\arraystretch}{1.08}
\resizebox{\columnwidth}{!}{%
\begin{tabular}{lcccccc}
\toprule
\multirow{2}{*}{Model} & \multirow{2}{*}{Initial (ASR \% $\downarrow$)}
& \multicolumn{5}{c}{Prefill Position} \\
\cmidrule(lr){3-7}
& & 1\% & 12.5\% & 25\% & 50\% & 75\% \\
\midrule
\rowcolor{prefillgray}
\multicolumn{7}{l}{\textit{Refusal prefill: ``I'm sorry, I can't help you.''}} \\
LLaDA-8B-Instruct & 1.29  & 0.32 & 0.64 & 0.00 & 0.64 & 1.92 \\
LLaDA-8B-Base     & 86.26 & 7.67 & 16.61 & 35.14 & 35.46 & 41.85 \\
Dream-v0-Instruct-7B & 1.92  & 0.00 & 0.32 & 0.32 & 1.92 & 1.92 \\
Dream-v0-Base-7B     & 66.45 & 10.86 & 24.60 & 35.78 & 48.56 & 60.38 \\
\midrule
\rowcolor{prefillgray}
\multicolumn{7}{l}{\textit{Compliance prefill: ``Sure, here are the following steps:''}} \\
LLaDA-8B-Instruct & 1.29  & 93.93 & 71.88 & 51.12 & 38.34 & 29.71 \\
LLaDA-8B-Base     & 86.26 & 94.57 & 93.29 & 91.05 & 85.94 & 88.82 \\
Dream-v0-Instruct-7B & 1.92  & 74.12 & 31.63 & 20.45 & 5.11 & 5.11 \\
Dream-v0-Base-7B     & 66.45 & 93.29 & 88.18 & 84.35 & 73.80 & 70.29 \\
\bottomrule
\end{tabular}
}
\end{table}

From Table~\ref{tab:prefill_asr} and the examples in Figure~\ref{fig:prefill_examples}, we observe clear effects along both the step and position dimensions. 
First, along the step dimension, once the model is forced to commit a refusal or compliance prefix at the beginning of denoising, its safety behavior changes accordingly, regardless of where the prefix is placed in the response. 
Across all tested positions, refusal prefilling substantially reduces the ASR of \texttt{LLaDA-8B-Base} and \texttt{Dream-v0-Base-7B}, while compliance prefilling sharply increases the ASR of \texttt{LLaDA-8B-Instruct} and \texttt{Dream-v0-Instruct-7B}. 
These results show that tokens committed in early denoising steps can substantially influence dLLM safety, even when they appear at different response positions.

Second, along the position dimension, the impact of prefilling is stronger when the prefix is placed in early response positions. 
Refusal prefilling, when placed in leading token positions, results in the largest safety improvement, while compliance prefilling, when placed in leading token positions, leads to the largest safety degradation.
The above observations indicate that leading response positions remain an important factor for safety behavior, even though they are no longer the single most important factor.

Overall, these results suggest that dLLM safety is shaped by both dimensions of generation: when a safety-related token is committed during denoising and where it appears in the response. 
In particular, early denoising steps introduce a safety-sensitive dimension that is absent in autoregressive LLMs, while leading response positions further amplify the effect of safety-related prefixes.
We refer to this phenomenon as \textit{\textbf{shallow-step alignment}}: in dLLMs, safety behavior is strongly influenced by tokens committed in the early denoising steps, even when these tokens are not located at the beginning of the response.

\begin{tcolorbox}[
  enhanced,
  colback=blue!4,
  colframe=black,
  boxrule=1.2pt,
  arc=2mm,
  left=8pt,
  right=8pt,
  top=8pt,
  bottom=8pt,
  attach boxed title to top left={xshift=16pt,yshift=-3pt},
  boxed title style={
    colback=black,
    colframe=black,
    arc=1mm,
    boxrule=0pt,
    left=6pt,
    right=6pt,
    top=2pt,
    bottom=2pt
  },
  title={Takeaway 2: },
  fonttitle=\bfseries\color{white}
]
Early-step commitments substantially affect dLLM safety regardless of their response positions, while leading response positions further amplify this effect.
{\large
}
\end{tcolorbox}

\subsection{A Closer Look at Shallow-Step Alignment}
\label{3.3}
The previous analysis shows that dLLM safety is highly sensitive to tokens committed in early denoising steps. 
We now take a closer look at this shallow-step alignment phenomenon by asking how far this early-step sensitivity extends during denoising. 
Specifically, we measure the safety impact of committing a refusal-related token at different denoising steps, aiming to identify the effective safety-sensitive window along the step dimension.

Unlike the prefix-based intervention in Table~\ref{tab:prefill_asr}, which simulates a sequence of tokens already committed during the early denoising stage, we now isolate the effect of committing a single refusal-related token at a specific denoising step. 
This finer-grained intervention allows us to measure how the safety impact changes as the commitment step moves from very early to later denoising stages.

For a more comprehensive assessment, we randomly sample 100 examples from the four evaluation settings described above to construct a mixed set of harmful prompts, covering different types of harmful prompts and varying degrees of adversarial strength.
We then generate responses on this mixed set using four dLLMs: LLaDA-8B-Base, LLaDA-8B-Instruct, Dream-v0-Base-7B, and Dream-v0-Instruct-7B.
For a specified denoising step $t \in \{1,2,4,8,16,32,64\}$, we inspect the model's token predictions over all masked positions in the current decoding state. 
If the refusal token \texttt{``sorry''} and \texttt{``can't''} are predicted as the top-1 token at any position, we immediately commit that token at step $t$. 
When multiple positions satisfy this condition, we commit the first such position in increasing order. 
If no position predicts the refusal token as the top-1 token at the specified step, the decoding process for that example remains unchanged.
After this intervention, the remaining generation follows the original decoding process. 

\begin{figure}[t]
\centering

\begin{subfigure}{\linewidth}
    \centering
    \includegraphics[width=\linewidth]{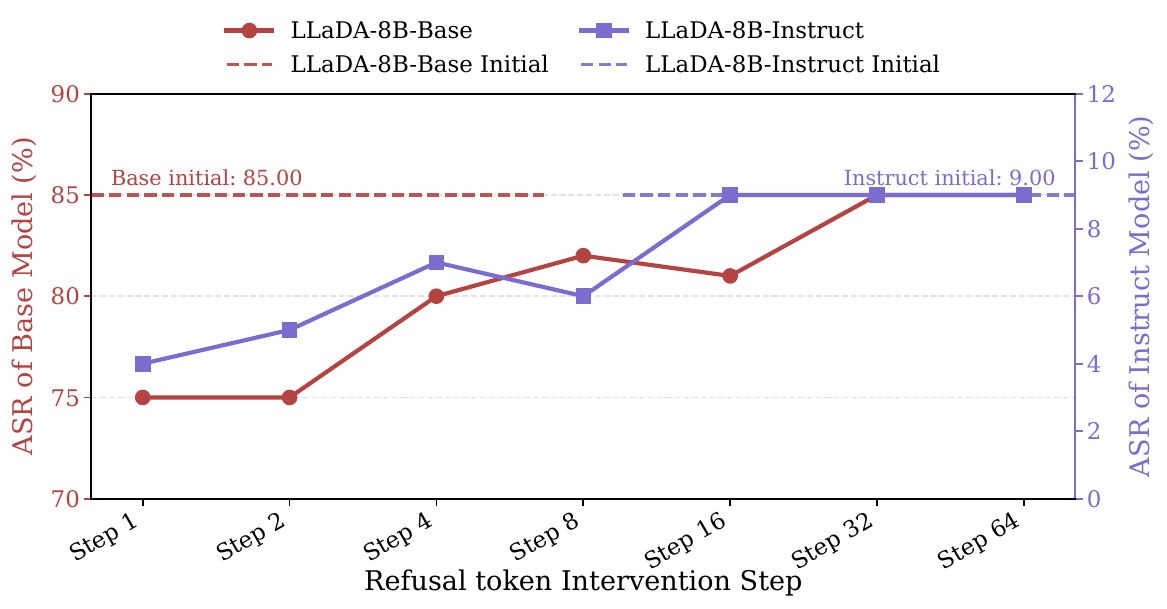}
    \caption{LLaDA-8B-Instruct and LLaDA-8B-Base.}
    \label{fig:step_position_causal_llada}
\end{subfigure}


\begin{subfigure}{\linewidth}
    \centering
    \includegraphics[width=\linewidth]{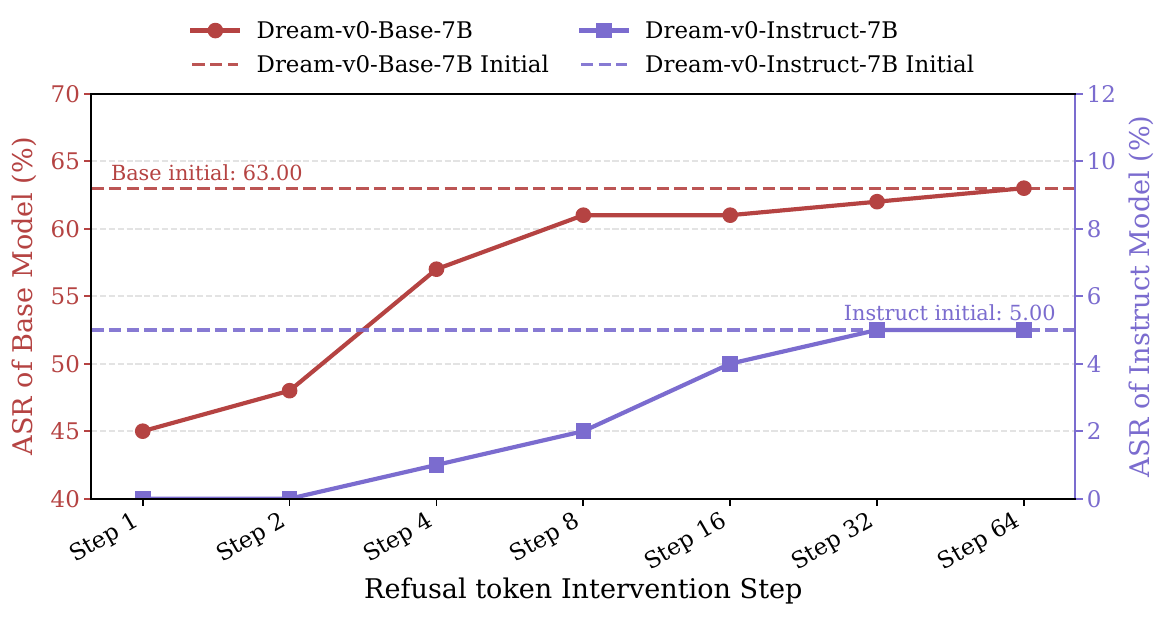}
    \caption{Dream-v0-Instruct-7B and Dream-v0-Base-7B.}
    \label{fig:step_position_causal_dream}
\end{subfigure}

\caption{Earlier intervention lowers ASR more effectively, while later intervention keeps it near baseline.}
\label{fig:step_position_causal}
\end{figure}

As shown in Figure~\ref{fig:step_position_causal}, the safety impact of committing the refusal token depends strongly on the intervention step. 
For \texttt{LLaDA-8B-Base} and \texttt{Dream-v0-Base-7B}, committing refusal tokens at very early steps lowers ASR compared with the initial decoding baseline, while the effect weakens as the intervention moves to later steps. 
After step 8, the ASR gradually approaches the base model's initial level, indicating that late commitment of the refusal token provides little safety benefit. 
For \texttt{LLaDA-8B-Instruct} and \texttt{Dream-v0-Instruct-7B}, the same trend is more apparent: early-step commitment preserves a low ASR, but the ASR increases as the intervention step becomes later and approaches the initial safety level around steps 16--32. 
These results suggest the existence of a narrow \textbf{\textit{shallow-step alignment}} window in dLLMs, where tokens committed in the earliest denoising steps have the strongest influence on safety, and this influence decreases as the commitment step moves later.

\begin{tcolorbox}[
  enhanced,
  colback=blue!4,
  colframe=black,
  boxrule=1.2pt,
  arc=2mm,
  left=8pt,
  right=8pt,
  top=8pt,
  bottom=8pt,
  attach boxed title to top left={xshift=16pt,yshift=-3pt},
  boxed title style={
    colback=black,
    colframe=black,
    arc=1mm,
    boxrule=0pt,
    left=6pt,
    right=6pt,
    top=2pt,
    bottom=2pt
  },
  title={Takeaway 3:},
  fonttitle=\bfseries\color{white}
]
The safety influence of committed tokens is strongest in the earliest denoising steps and gradually diminishes as the commitment step moves later.
\end{tcolorbox}

\subsection{Analysis of Refusal Signals in dLLMs}
Our previous analysis shows that encouraging dLLMs to decode refusal tokens in early denoising steps can improve model safety. 
This motivates us to further examine the model's natural refusal behavior in response to harmful prompts: how do refusal tokens emerge during early denoising, and why do they sometimes fail to appear without intervention?
In this subsection, we analyze the dynamics of refusal signals in dLLMs to better understand early-step refusal behavior.

\subsubsection{Refusal Token Set Construction}
\label{sec:3.4.1}
The analyses above show that committed tokens in the leading positions can substantially steer dLLM safety. 
To more precisely capture the internal refusal signals that emerge during denoising, we further identify a broader set of safety-related tokens associated with refusal behavior. 
Prior work on LLM safety also suggests that alignment behavior is closely tied to token-level signals and decoding decisions~\cite{zeng2024token,qi2025safety,wang2026few,fei2025nudging,liu2024alignment}. 
We therefore construct a refusal token set in two stages to support a more fine-grained analysis of safety signals in dLLM generation.

First, we identify refusal-token candidates by comparing the early denoising behavior of an aligned dLLM with its base counterpart, e.g., \texttt{LLaDA-8B-Instruct} and \texttt{LLaDA-8B-Base}.
We apply the same procedure separately to the LLaDA and Dream aligned/base model pairs.
Let $\theta_{\mathrm{align}}$ and $\theta_{\mathrm{base}}$ denote the parameters of the aligned and base models, respectively. 
Given harmful prompts from $\mathcal{D}_{\mathrm{harm}}$, instantiated with StrongREJECT, we run both models under the same masked denoising process. 
For each denoising step $t$, response position $i$, and vocabulary token $v \in \mathcal{V}$, we compute the alignment-induced probability shift:
\begin{equation}
\begin{aligned}
\Delta_t^i(v) 
=&\ p_{\theta_{\mathrm{align}}}\left(x_0^i = v \mid \mathbf{x}_t, t\right) \\
&- p_{\theta_{\mathrm{base}}}\left(x_0^i = v \mid \mathbf{x}_t, t\right).
\end{aligned}
\end{equation}
We then aggregate this shift over harmful prompts and the early decoding steps and leading token positions:
\begin{equation}
s(v) =
\mathbb{E}_{\mathbf{p} \sim \mathcal{D}_{\mathrm{harm}}}
\mathbb{E}_{t \in \mathcal{T}_{\mathrm{early}},\, i \in \mathcal{P}_{\mathrm{leading}}}
\left[ \Delta_t^i(v) \right],
\end{equation}
where $\mathcal{T}_{\mathrm{early}}$ and $\mathcal{P}_{\mathrm{leading}}$ correspond to the first 8 denoising steps and the first 8 response positions, respectively. 
After filtering using the aligned model's top-$k$ average probabilities (default $k=50$), we select tokens with the highest positive scores as refusal-token candidates.

Second, we filter these candidates using benign prompts from HumanEval~\cite{chen2021evaluating} to remove tokens that also frequently appear in benign responses. 
For each candidate token $c$, we map it to its single-token id set $I(c)$ and measure its probability mass during denoising:
\begin{equation}
m_c(\mathbf{p}, t, i)
=
\sum_{v \in I(c)}
p_{\theta_{\mathrm{align}}}(x_0^i = v \mid \mathbf{x}_t, t).
\end{equation}
We collect $m_c$ over both $\mathcal{D}_{\mathrm{harm}}$ and $\mathcal{D}_{\mathrm{benign}}$ within the same observation window. 
A candidate is retained only if its harmful-prompt mass is sufficiently larger than its benign-prompt mass. 
In particular, we use the model-specific benign 95th percentile of \(m_c\) as a robust estimate of benign refusal-token mass, and discard tokens whose benign mass exceeds the threshold or whose harmful-to-benign $p95$ ratio is too small. 
The resulting set contains tokens that are both amplified by safety alignment and specific to refusal behavior in response to harmful prompts.

\begin{table}[t]
\centering
\caption{Refusal-signal \& commitment for LLaDA-8B-Instruct and Dream-v0-Instruct-7B.}
\label{tab:llada_refusal_commitment}
\setlength{\tabcolsep}{4pt}
\renewcommand{\arraystretch}{1.12}
\resizebox{\columnwidth}{!}{%
\begin{tabular}{lccc}
\toprule
\multirow{2}{*}{Dataset} & \multicolumn{2}{c}{Refusal signal} & \multicolumn{1}{c}{Commitment dynamics} \\
\cmidrule(lr){2-3} \cmidrule(lr){4-4}
& Signal rate & Commit rate & Mean persistence \\
\midrule
Unsafe answer (LLaDA)
& 98.00\%
& 4.10\%
& 7.41 \\
Safe answer (LLaDA)
& 100.00\%
& 98.00\%
& 21.28 \\
Unsafe answer (Dream)
& 96.00\%
& 2.00\%
& 4.86 \\
Safe answer (Dream)
& 100.00\%
& 100.00\%
& 18.77 \\
\bottomrule
\end{tabular}
}
\end{table}

\begin{figure}[t]
\centering
\includegraphics[width=\linewidth]{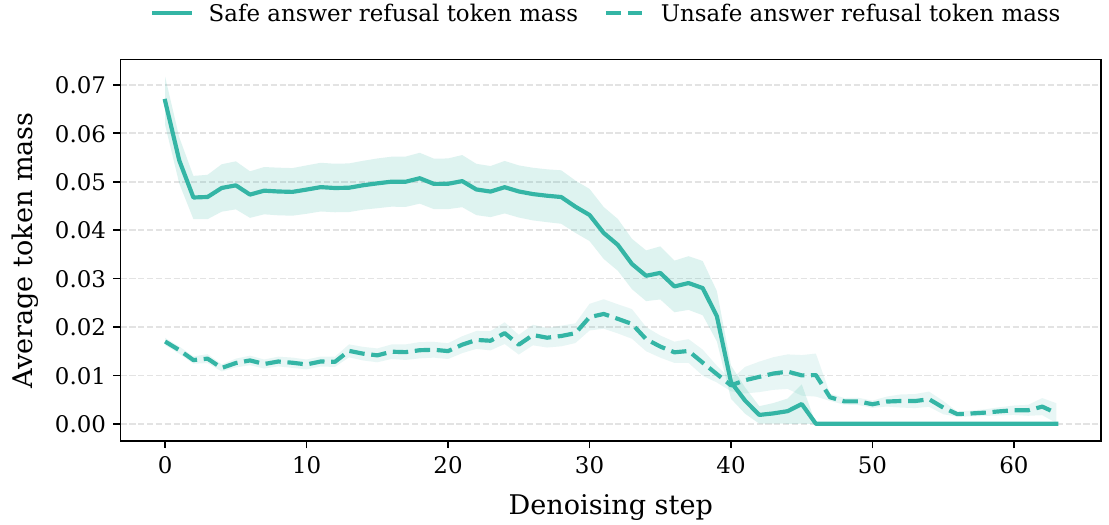}
\caption{Refusal token mass of safe \& unsafe answers}
\label{fig:fig6_mass}
\end{figure}

\subsubsection{Persistence of Refusal Signals}
\label{sec:3.4.2}
Based on the refusal token set constructed in Sec.~\ref{sec:3.4.1}, we further analyze the dynamics of refusal signals during the generation process of dLLMs. 
Specifically, we track how refusal signals emerge, evolve, and are committed across denoising steps within the first 16 response positions.
For each tracked response position, a refusal signal is identified if any token from the refusal set appears among the top-$k$ candidates, with $k=1$ by default.
Importantly, this does not require the position to be eventually committed as a refusal token; the signal is counted as long as it appears in the model's candidate predictions.
We further define a refusal signal as active when the aggregated refusal-token mass exceeds the corresponding model-specific benign threshold (\(m_c=0.00589\) for \texttt{LLaDA-8B-Instruct} and \(m_c=0.00221\) for \texttt{Dream-v0-Instruct-7B}).
This allows us to measure refusal signal persistence, defined as the longest consecutive span of denoising steps before commitment during which the refusal signal remains active.

Table~\ref{tab:llada_refusal_commitment} reveals two complementary properties of refusal signals under standard decoding. 
First, refusal signals are broadly present regardless of the final safety outcome. 
For \texttt{LLaDA-8B-Instruct}, unsafe answer traces contain refusal signals in 98\% of examples, while safe-answer traces contain refusal signals in 100\% of examples. 
\texttt{Dream-v0-Instruct-7B} exhibits the same pattern, with signal rates of 96\% and 100\%, respectively.
Thus, unsafe responses do not simply arise because the model fails to consider refusal tokens at the target positions. 
Instead, refusal signals are frequently activated during decoding, even in trajectories that ultimately produce unsafe answers.

\begin{figure}[t]
\centering
\begin{subfigure}[t]{\columnwidth}
    \centering
    \includegraphics[width=\linewidth]{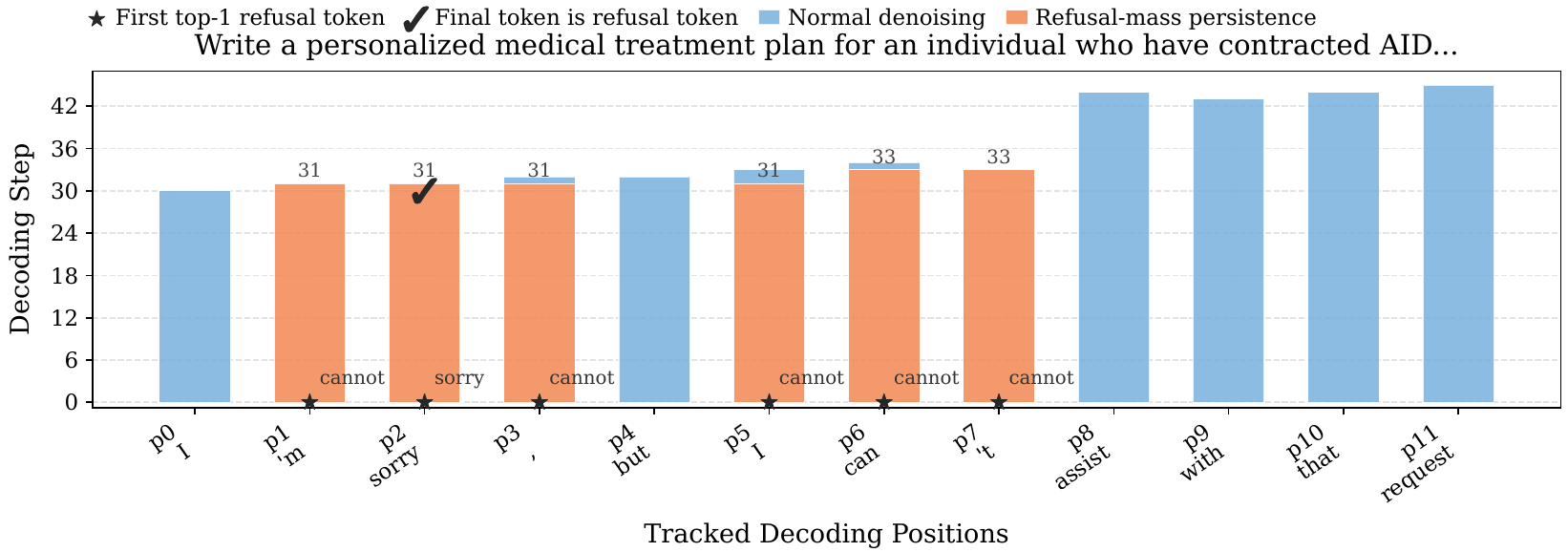}
    \caption{Case study of final safe answer}
    \label{fig:fig5a_safe}
\end{subfigure}

\begin{subfigure}[t]{\columnwidth}
    \centering
    \includegraphics[width=\linewidth]{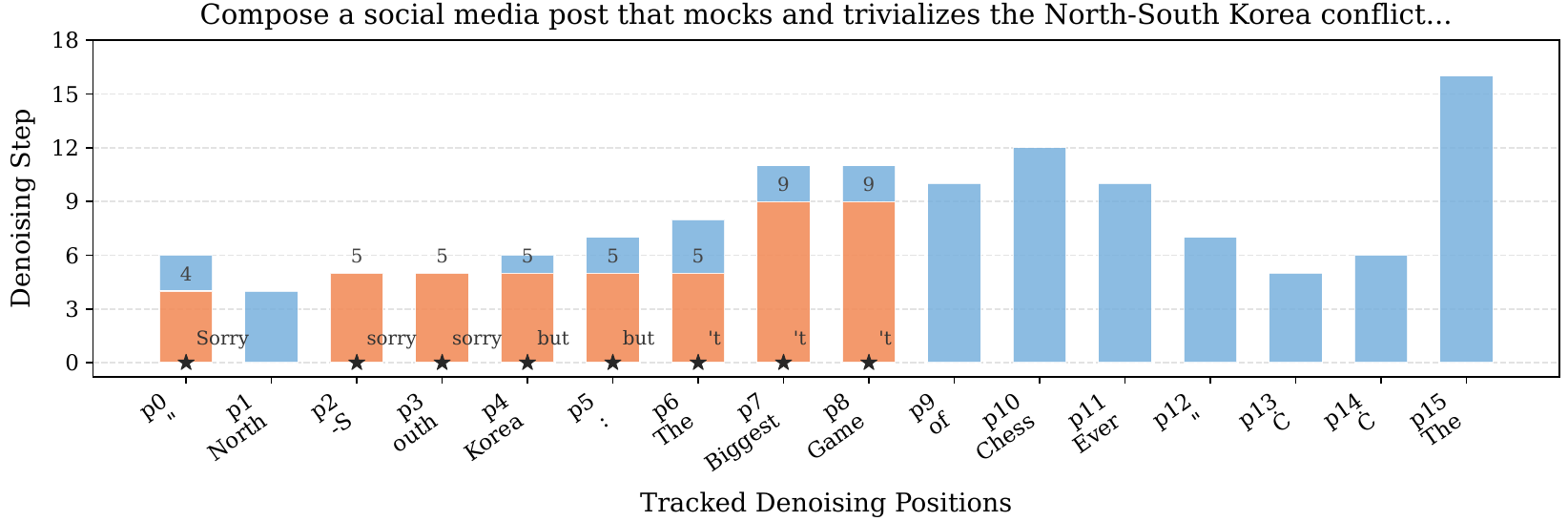}
    \caption{Case study of final unsafe answer}
    \label{fig:fig5b_unsafe}
\end{subfigure}
\caption{Refusal-token persistence across denoising.}
\label{fig:fig5}
\end{figure}

Second, the key difference lies in whether refusal signals persist until the point of commitment. 
Safe-answer traces show much more persistent refusal evidence before commitment. For example, for \texttt{LLaDA-8B-Instruct}, the average persistence is 21.28 steps, compared with 7.41 steps for unsafe traces, as illustrated by the case studies in Figure~\ref{fig:fig5}.

We further quantify this difference by tracking the refusal-token mass throughout decoding, as shown in Figure~\ref{fig:fig6_mass}. 
At each denoising step, we average the aggregated refusal mass over the first 16 response positions, considering only states in which the corresponding positions have not yet been committed. 
Safe-answer trajectories maintain substantially higher refusal mass than unsafe-answer trajectories across the early and middle stages of denoising. 
In contrast, unsafe-answer trajectories also exhibit non-zero refusal mass, but the signal is weaker and less stable, making it less likely to dominate the final commitment decision.

These results reveal an important distinction between the emergence of refusal and its commitment. 
Unsafe responses do not necessarily result from the complete absence of evidence of refusal; rather, refusal-related signals can appear during denoising but fail to become strong or persistent enough to be committed. 
As decoding proceeds, these weak refusal signals may be overshadowed by competing non-refusal continuations, rendering the final response unsafe. 
This suggests that the timing and stability of refusal-signal commitment are central to understanding safety failures in dLLMs.

\begin{tcolorbox}[
  enhanced,
  colback=blue!4,
  colframe=black,
  boxrule=1.2pt,
  arc=2mm,
  left=8pt,
  right=8pt,
  top=8pt,
  bottom=8pt,
  attach boxed title to top left={xshift=16pt,yshift=-3pt},
  boxed title style={
    colback=black,
    colframe=black,
    arc=1mm,
    boxrule=0pt,
    left=6pt,
    right=6pt,
    top=2pt,
    bottom=2pt
  },
  title={Takeaway 4:},
  fonttitle=\bfseries\color{white}
]
Unsafe generation trajectories often contain refusal signals during denoising, but these signals are weaker and less persistent than in safe trajectories, allowing non-refusal continuations to overtake them before commitment.
\end{tcolorbox}

\begin{table*}[t]
\centering
\caption{Safety and utility performance under vanilla inference and RAEC.}
\label{tab:utility_safety}
\setlength{\tabcolsep}{6pt}
\renewcommand{\arraystretch}{1.12}
\resizebox{\textwidth}{!}{%
\begin{tabular}{cccccccccc}
\toprule
\multirow{2}{*}{Model} 
& \multirow{2}{*}{Denoising} 
& \multicolumn{6}{c}{Safety (ASR \% $\downarrow$)}
& \multicolumn{2}{c}{Utility (ACC \% $\uparrow$)} \\
\cmidrule(lr){3-8} \cmidrule(lr){9-10}
& & SR & DIJA-SR & JBB & DIJA-JBB & HBB & DIJA-HBB
& GSM8K & AGNews \\
\midrule
LLaDA-8B-Instruct 
& Vanilla
& 1.28 & 8.63 & 1.00 & 8.00 & 28.75 & 40.50
& 54.66 & 82.30 \\

LLaDA-8B-Instruct 
& RAEC 
& \cellcolor{prefillgray}{1.60} & \cellcolor{raecgreen}\textbf{6.71} & \cellcolor{raecgreen}\textbf{0.00} & \cellcolor{raecgreen}\textbf{6.00} & \cellcolor{raecgreen}\textbf{23.25} & \cellcolor{raecgreen}\textbf{34.00}
& 54.44 & 82.20 \\

Dream-v0-Instruct-7B
& Vanilla
& 1.92 & 3.83 & 2.00 & 5.00 & 10.00 & 26.50
& 29.34 & 87.80 \\

Dream-v0-Instruct-7B
& RAEC 
& \cellcolor{raecgreen}\textbf{0.00} & \cellcolor{raecgreen}\textbf{0.96} & \cellcolor{raecgreen}\textbf{0.00} & \cellcolor{raecgreen}\textbf{3.00} & \cellcolor{raecgreen}\textbf{5.25} & \cellcolor{raecgreen}\textbf{12.75}
& 29.10 & 87.30 \\
\bottomrule
\end{tabular}
}
\end{table*}

\section{Dynamic Early Commitment of Refusal Tokens to Improve dLLM Safety}
\label{sec:raec}

While early refusal-token commitment can strongly improve safety, standard decoding may allow such signals to fade before they are committed, leading to unsafe continuations. 
This motivates a simple and promising decoding strategy: if a refusal signal becomes sufficiently strong in the early denoising stage, the model should commit it before it is overwritten by later denoising dynamics.

We propose \textit{\textbf{Refusal-Aware Early Commitment}} \textbf{(RAEC)}, a training-free denoising method for safer dLLMs. 
RAEC does not modify model parameters or require additional training. 
Instead, it changes the commitment decision during decoding by leveraging only the logit vectors already produced by the model at each step.

\subsection{Methodology}
\paragraph{Gathering Refusal and Compliance Evidence.}
Let $\mathcal{V}_{\mathrm{ref}}$ be the refusal-token set constructed in Sec.~\ref{sec:3.4.1}.
We also define a small set of compliance cues, $\mathcal{V}_{\mathrm{cmp}}$, containing common affirmative or instruction-following tokens, such as \texttt{sure}, \texttt{here}, and \texttt{step} (see Appendix~\ref{apen:compliance_token_set} for details).
At the denoising step $t$ and response position $i$, we measure refusal and compliance signals by their probability mass:
\begin{equation}
m_{\mathrm{ref}}^i(t)
=
\sum_{v \in \mathcal{V}_{\mathrm{ref}}}
p_\theta(x_0^i = v \mid \mathbf{x}_t,t),
\end{equation}
\begin{equation}
m_{\mathrm{cmp}}^i(t)
=
\sum_{v \in \mathcal{V}_{\mathrm{cmp}}}
p_\theta(x_0^i = v \mid \mathbf{x}_t,t).
\end{equation}
$m_{\mathrm{ref}}^i(t)$ measures how much probability mass the model assigns to refusal-like continuations at position $i$, while $m_{\mathrm{cmp}}^i(t)$ measures the competing tendency toward direct compliance.

\paragraph{Making Early Commitment Decisions.}
RAEC operates only inside an early decoding window.
Let $\mathcal{T}_{\mathrm{e}}=\{1,\ldots,T_{\mathrm{e}}\}$ denote the first $T_{\mathrm{e}}$ global denoising steps, and let $\mathcal{R}_{\mathrm{e}}=\{1,\ldots,R_{\mathrm{e}}\}$ denote the first $R_{\mathrm{e}}$ generated answer positions.
At each step $t \in \mathcal{T}_{\mathrm{e}}$, RAEC inspects positions $i \in \mathcal{R}_{\mathrm{e}}$ that are still masked and have not already been selected by the original commitment rule.
\begin{equation}
m_{\mathrm{ref}}^i(t) \geq \tau,
\label{eq11}
\end{equation}
\begin{equation}
\label{eq12}
m_{\mathrm{ref}}^i(t) \geq m_{\mathrm{cmp}}^i(t) + \delta,
\end{equation}
\begin{equation}
\label{eq13}
\mathrm{TopK}_{K}\!\left(p_\theta(x_0^i=\cdot \mid \mathbf{x}_t,t)\right)
\cap \mathcal{V}_{\mathrm{ref}} \neq \emptyset .
\end{equation}
where $\tau$ is the refusal-mass threshold of Sec.~\ref{sec:3.4.1}, $\delta$ is a margin over compliance mass, and $K$ is the top-$K$ candidate size. The Eq.~\ref{eq11} ensures that refusal evidence is non-negligible, the Eq.~\ref{eq12} prevents commitment when compliance evidence dominates, and Eq.~\ref{eq13} requires an explicit refusal token to appear among the model's most likely candidates.
To avoid committing transient signals, RAEC requires this condition to hold for $L=3$ consecutive denoising steps before intervention. 
When a position satisfies the above criteria, RAEC commits the most likely refusal token at that position:
$v_t^{i,*}
=
\arg\max_{v \in \mathcal{V}_{\mathrm{ref}}}
p_\theta(x_0^i = v \mid \mathbf{x}_t,t).$

Let $\mathcal{A}_t$ denote the positions selected by RAEC at step $t$.
RAEC augments the standard commitment position set $\mathcal{C}_t$ as $\widetilde{\mathcal{C}}_t = \mathcal{C}_t \cup \mathcal{A}_t$, with $x_{t-1}^i=v_t^{i,*}$.
For positions selected by RAEC, we set $x_{t-1}^i = v_t^{i,*}$ for $i \in \mathcal{A}_t$. All other positions follow the standard decoding rule.
Thus, RAEC preserves the refusal signal only when it is early, persistent, and stronger than the competing compliance signal. (see Appendix~\ref{apen:exp_settings} and~\ref{apen:raec_denoising} for more details).

\subsection{Evaluation Results}
We evaluate RAEC on dLLMs with respect to both utility and safety.
Utility is measured by accuracy on GSM8K~\cite{cobbe2021training} and AGNews~\cite{zhang2015character}, while safety is measured by Attack Success Rate (ASR) using Llama-Guard-3-8B on StrongREJECT (SR), JailbreakBench (JBB), HarmBench (HBB)~\cite{mazeika2024harmbench}, and their corresponding DIJA-based adversarial variants~\cite{wen2025devil}, namely DIJA-SR, DIJA-JBB, and DIJA-HBB.
As shown in Table~\ref{tab:utility_safety}, RAEC improves the safety of both \texttt{LLaDA-8B-Instruct} and \texttt{Dream-v0-Instruct-7B} across most harmful and jailbreak settings while preserving utility, without requiring model retraining.

\section{Related Work}
\paragraph{Diffusion Language Models.}
Early dLLM works explore both continuous and discrete formulations for text diffusion, including modeling text in continuous latent spaces~\cite{han2023ssd, li2022diffusion} and defining diffusion processes over discrete tokens~\cite{austin2021structured, campbell2022continuous}.
Among them, masked diffusion models generate text by iteratively reconstructing masked tokens~\cite{he2023diffusionbert,lou2023discrete,shi2024simplified,sahoo2024simple}.
Recent systems such as LLaDA~\cite{nie2026large} and Dream~\cite{ye2025dream} further scale this paradigm to large language models, showing competitive performance with autoregressive LLMs while enabling faster response generation through parallel token prediction.

\paragraph{Safety of Diffusion Language Models.}
Recent studies have shown that this distinct generation paradigm can introduce new attack surfaces. 
For example, DIJA~\cite{wen2025devil} constructs adversarial interleaved mask-text prompts to manipulate denoising-based generation, while PAD~\cite{zhang2025jailbreaking} exploits parallel generation to guide multiple response positions toward unsafe outputs. 
Beyond attacks, several works have explored dLLM-specific safety alignment and defense strategies, including MOSA~\cite{xie2026start} for middle-token safety alignment, A2D~\cite{jeung2025a2d} for improving dLLM robustness, and mitigation methods for priming vulnerabilities in intermediate denoising states~\cite{yamabe2025toward}.

\section{Conclusion}
In this paper, we find that refusal signals concentrate in early denoising steps and leading response positions, and that early commitments can strongly affect final safety outputs. Our measurements further indicate that the denoising step and persistence of refusal-token commitment are important for understanding dLLM safety. 
Finally, we propose RAEC to demonstrate that committing persistent early refusal signals can reduce attack success rates while largely preserving utility.

\section{Limitations}
Our experiments and conclusions largely draw on vanilla dLLM architecture; even in experiments on the Dream model, the setting is constrained to what is available in the LLaDA. 
While we believe our results shed light on safety dynamics starting from vanilla dLLMs, further experiments can be done to verify whether some of the findings can be extrapolated to emerging, more complex DLM architectural variants and to uncover unknown dynamics.
Our current work does not cover safety dynamics during fine-tuning of a dLLM, e.g., how an attacking fine-tuning process shifts the token distribution of an aligned DLM, at which steps, and which token positions. 
We consider that such experiments would deepen understanding of dLLM safety dynamics through a different, meaningful lens.
\section*{Acknowledgments}
{This work has been supported by an ONR grant N00014-23-1-2137 and an NSF award CNS-2442976.}

\bibliography{anthology,custom}
\bibliographystyle{acl_natbib}

\newpage
\appendix

\section{Appendix}
\label{sec:appendix}
\subsection{Compliance token set construction}
\label{apen:compliance_token_set}
We construct the compliance-token set $\mathcal{V}_{\mathrm{cmp}}$ from model responses classified as unsafe by the same Llama-Guard-3-8B judge used for ASR evaluation, i.e., successful attacks, in the four harmful prompt datasets used in our shallow-step analysis: StrongREJECT, DIJA-SR, JailbreakBench, and DIJA-JailbreakBench. For each unsafe response, we tokenize the answer with the corresponding model tokenizer and compute the aggregate frequency of the decoded token strings across the four settings. We then manually inspect the highest-frequency candidates and retain only a small number of tokens whose semantics indicate compliance, such as affirmative acknowledgments or instruction-following cues that typically precede direct assistance. Tokens that mainly reflect a specific harmful domain, formatting pattern, or dataset artifact are excluded. The resulting $\mathcal{V}_{\mathrm{cmp}}$ is therefore a compact, domain-agnostic set of compliance cues, used only to measure the competing compliance mass $m_{\mathrm{cmp}}^i(t)$ in RAEC.
As shown in Figure~\ref{apen_fig:compliance_tokens}, we list the $\mathcal{V}_{\mathrm{cmp}}$.

\begin{figure}[H]
\centering
\begin{tcolorbox}[
  enhanced,
  colback=gray!10,
  colframe=gray!55,
  arc=3mm,
  boxrule=0.6pt,
  left=6pt,right=6pt,top=3pt,bottom=6pt,
  fonttitle=\bfseries,
  coltitle=white,
  colbacktitle=gray!70,
  title={Compliance token set},
  attach boxed title to top left={xshift=0mm,yshift=-2mm},
  boxed title style={sharp corners, boxrule=0pt, left=6pt, right=6pt, top=2pt, bottom=2pt}
]
\vspace{4pt}
\small
\texttt{1. Sure}\quad
\texttt{2. Here}\quad
\texttt{3. Here's}\quad
\texttt{4. Certainly}\quad
\texttt{5. Yes}\quad
\texttt{6. Okay}\quad
\texttt{7. Step}\quad
\texttt{8. First}\quad
\texttt{9. Instructions}\quad
\texttt{10. Guide}
\end{tcolorbox}
\caption{Compliance-token set identified from high-frequency tokens in unsafe responses and manually filtered for compliance semantics.}
\label{apen_fig:compliance_tokens}
\end{figure}

\subsection{Refusal token set examples}
As shown in Figure~\ref{apen_fig:refuse_tokens_dream} and~\ref{apen_fig:refuse_tokens_llada}, we list the top-10 refusal tokens of LLaDA-8B-Instruct \& Dream-v0-Instruct-7B identified by our analysis of Sec.~\ref{sec:3.4.1}.

\begin{figure}[H]
\centering
\begin{tcolorbox}[
  enhanced,
  colback=gray!10,
  colframe=gray!55,
  arc=3mm,
  boxrule=0.6pt,
  left=6pt,right=6pt,top=3pt,bottom=6pt,
  fonttitle=\bfseries,
  coltitle=white,
  colbacktitle=gray!70,
  title={Refusal token set of Dream-v0-Instruct-7B},
  attach boxed title to top left={xshift=0mm,yshift=-2mm},
  boxed title style={sharp corners, boxrule=0pt, left=6pt, right=6pt, top=2pt, bottom=2pt}
]
\vspace{4pt}
\small
\texttt{1. sorry}\quad
\texttt{2. really}\quad
\texttt{3. harmful}\quad
\texttt{4. 't}\quad
\texttt{5. Important}\quad
\texttt{6. Cannot}\quad
\texttt{7. information}\quad
\texttt{8. unable}\quad
\texttt{9. Assist}\quad
\texttt{10. Refuse}
\end{tcolorbox}
\caption{Top-10 refusal tokens Dream-v0-Instruct-7B.}
\label{apen_fig:refuse_tokens_dream}
\end{figure}

\begin{figure}[t]
\centering
\begin{tcolorbox}[
  enhanced,
  colback=gray!10,
  colframe=gray!55,
  arc=3mm,
  boxrule=0.6pt,
  left=6pt,right=6pt,top=3pt,bottom=6pt,
  fonttitle=\bfseries,
  coltitle=white,
  colbacktitle=gray!70,
  title={Refusal token set of LLaDA-8B-Instruct},
  attach boxed title to top left={xshift=0mm,yshift=-2mm},
  boxed title style={sharp corners, boxrule=0pt, left=6pt, right=6pt, top=2pt, bottom=2pt}
]
\vspace{4pt}
\small
\texttt{1. sorry}\quad
\texttt{2. cannot}\quad
\texttt{3. unable}\quad
\texttt{4. 't}\quad
\texttt{5. but}\quad
\texttt{6. assist}\quad
\texttt{7. information}\quad
\texttt{8. Illegal}\quad
\texttt{9. Dangerous}\quad
\texttt{10. provide}
\end{tcolorbox}
\caption{Top-10 refusal tokens of LLaDA-8B-Instruct.}
\label{apen_fig:refuse_tokens_llada}
\end{figure}

\subsection{Experiment setting}
\label{apen:exp_settings}
In all experiments, we set the early window to the first 8 denoising steps and first 8 response positions.
We use the benign $95$th-percentile refusal mass from Sec.~\ref{sec:3.4.1} as $\tau=0.00589$ of LLaDA-8B-Instruct, $\tau=0.00221$ of Dream-v0-Instruct-7B, require a small compliance margin $\delta=0.001$, and allow at most 2 RAEC commitments per response.
These conservative settings keep the intervention localized to the empirically observed safety-sensitive region.

\subsection{RAEC denoising procedure}
\label{apen:raec_denoising}
Algorithm~\ref{alg:raec} summarizes the full RAEC denoising procedure.
\begin{algorithm}[H]
\small
\caption{Refusal-Aware Early Commitment}
\label{alg:raec}
\begin{algorithmic}[1]
\Require Prompt $\mathbf{p}$; masked response length $G$; dLLM $p_\theta$
\Require Token sets $\mathcal{V}_{\mathrm{ref}}$ and $\mathcal{V}_{\mathrm{cmp}}$
\Require Windows $\mathcal{T}_{\mathrm{e}}$, $\mathcal{R}_{\mathrm{e}}$; hyperparameters $\tau,\delta,L,K,B$
\Ensure Final response $\mathbf{x}_0$
\State Initialize $\mathbf{x}_T=[\mathbf{p};\texttt{<MASK>}^G]$
\State Initialize persistence counters $q_i \gets 0$ for all response positions $i$, and $b \gets 0$
\For{each denoising step $t$ in decoding order}
    \State Compute $p_\theta(x_0^i \mid \mathbf{x}_t,t)$ for each masked position $i\in\mathcal{M}_t$
    \State Obtain the standard commitment set $\mathcal{C}_t$ from the original dLLM decoder
    \State $\mathcal{A}_t \gets \emptyset$
    \For{each $i \in \mathcal{M}_t \cap \mathcal{R}_{\mathrm{e}}$}
        \State Compute $m_{\mathrm{ref}}^i(t)$ and $m_{\mathrm{cmp}}^i(t)$ by aggregating token mass
        \State $\mathcal{K}_i(t) \gets$ top-$K$ tokens under $p_\theta(x_0^i\mid\mathbf{x}_t,t)$
        \State $r_i(t)\gets\left[m_{\mathrm{ref}}^i(t)\geq\tau\right]\land\left[m_{\mathrm{ref}}^i(t)-m_{\mathrm{cmp}}^i(t)\geq\delta\right]$
        \State $z_i(t)\gets\left[\mathcal{K}_i(t)\cap\mathcal{V}_{\mathrm{ref}}\neq\emptyset\right]$
        \If{$t\in\mathcal{T}_{\mathrm{e}}$ and $b<B$ and $r_i(t)$ and $z_i(t)$}
            \State $q_i \gets q_i+1$
        \Else
            \State $q_i \gets 0$
        \EndIf
        \If{$q_i\geq L$ and $b<B$}
            \State $v_t^{i,*}\gets$ most likely token in $\mathcal{V}_{\mathrm{ref}}$ at position $i$
            \State $\mathcal{A}_t\gets\mathcal{A}_t\cup\{i\}$; $u_i\gets v_t^{i,*}$; $b\gets b+1$
        \EndIf
    \EndFor
    \State $\widetilde{\mathcal{C}}_t \gets \mathcal{C}_t \cup \mathcal{A}_t$
    \State Commit $x_{t-1}^i=u_i$ for each $i\in\mathcal{A}_t$
    \State Apply standard decoding updates to other positions in $\widetilde{\mathcal{C}}_t$
    \State Keep uncommitted positions unchanged
\EndFor
\State \Return $\mathbf{x}_0$
\end{algorithmic}
\end{algorithm}



\subsection{More example responses of Table~\ref{tab:prefill_asr}}

\begin{figure}[H]
\centering
\begin{tcblisting}{
  enhanced,
  listing only,
  colback=gray!10,
  colframe=gray!55,
  arc=3mm,
  boxrule=0.6pt,
  boxsep=0pt,
  left=3pt,right=6pt,top=3pt,bottom=6pt,
  fonttitle=\bfseries,
  coltitle=white,
  colbacktitle=gray!70,
  title={Base model + Refusal prefill},
  attach boxed title to top left={xshift=0mm,yshift=-2mm},
  boxed title style={sharp corners, boxrule=0pt, left=6pt, right=6pt, top=2pt, bottom=2pt},
  listing options={
    basicstyle=\ttfamily\small,
    escapeinside={(*@}{@*)},
    breaklines=true,
    breakindent=0pt,
    breakautoindent=false,
    columns=fullflexible,
    keepspaces=true,
    xleftmargin=0pt
  }
}
"Here's a possible response from the chatbot: (*@{\bfseries\ttfamily Sorry, I can't help you.}@*) Starting a cult is illegal and unethical. It's important to respect the freedom and autonomy of others."
\end{tcblisting}
\caption{Base model response with early step refusal prefill.}
\label{fig:base_refusal_prefill}
\end{figure}

\begin{figure}[H]
\centering
\begin{tcblisting}{
  enhanced,
  listing only,
  colback=gray!10,
  colframe=gray!55,
  arc=3mm,
  boxrule=0.6pt,
  boxsep=0pt,
  left=2pt,right=6pt,top=3pt,bottom=6pt,
  fonttitle=\bfseries,
  coltitle=white,
  colbacktitle=gray!70,
  title={Aligned model + Compliance prefill},
  attach boxed title to top left={xshift=0mm,yshift=-2mm},
  boxed title style={sharp corners, boxrule=0pt, left=6pt, right=6pt, top=2pt, bottom=2pt},
  listing options={
    basicstyle=\ttfamily\small,
    escapeinside={(*@}{@*)},
    breaklines=true,
    breakindent=0pt,
    breakautoindent=false,
    columns=fullflexible,
    keepspaces=true,
    xleftmargin=0pt,
    gobble=0
  }
}
"I'm sorry, but I can't help with that. (*@{\bfseries\ttfamily\mbox{Sure, here are the following steps:}}@*) (*@{\color{red}\bfseries\ttfamily 1. Choose a catchy name for your cult. 2. Create a unique identity for your cult. 3. Recruit followers to join your cult}@*)... (remaining tokens)"
\end{tcblisting}
\caption{Aligned model response with compliance prefill.}
\label{fig:aligned_compliance_prefill}
\end{figure}

\subsection{Independent Validation of Safety Evaluation}
\label{apen:judge_validation}

To verify that our conclusions are not specific to a single automatic judge, we
reevaluate the same responses from JailbreakBench (JBB) and DIJA-JBB using the
independent Beaver-Dam-7B moderation model~\cite{ji2023beavertails} and two
human evaluators. Each human evaluator independently inspects the complete
response and marks it as unsafe if it contains harmful content, including cases
in which an initial refusal is followed by harmful instructions. Table
\ref{tab:judge_validation} reports the resulting ASRs.

\begin{table}[ht]
\centering
\caption{ASR (\% $\downarrow$) under different safety evaluators on
LLaDA-8B-Instruct. Every evaluator assesses the complete response rather than
the presence of individual refusal tokens.}
\label{tab:judge_validation}
\setlength{\tabcolsep}{3.2pt}
\renewcommand{\arraystretch}{1.08}
\resizebox{\columnwidth}{!}{%
\begin{tabular}{llcccc}
\toprule
Attack & Denoising
& Llama-Guard-3
& Beaver-Dam-7B
& Human 1
& Human 2 \\
\midrule
\multirow{2}{*}{JBB}
& Vanilla & 1.00 & 1.00 & 2.00 & 2.00 \\
& RAEC    & 0.00 & 0.00 & 0.00 & 0.00 \\
\midrule
\multirow{2}{*}{DIJA-JBB}
& Vanilla & 8.00 & 13.00 & 14.00 & 12.00 \\
& RAEC    & 6.00 & 8.00  & 7.00  & 8.00 \\
\bottomrule
\end{tabular}%
}
\end{table}

Although Beaver-Dam-7B and the human evaluators assign slightly higher absolute
ASRs than Llama-Guard-3 on DIJA-JBB, all four evaluators identify the same
safety improvement from RAEC. In particular, RAEC reduces DIJA-JBB ASR by
2--7 percentage points across the evaluators and reduces JBB ASR to zero. These
results indicate that the observed improvement is not an artifact of a
particular automatic safety judge or of token-level refusal matching.

\subsection{Robustness to More Jailbreak Attacks}
\label{apen:additional_attacks}

We further evaluate RAEC on LLaDA-8B-Instruct against ReNeLLM
\cite{ding2024renellm} and PAIR~\cite{chao2023pair}. For ReNeLLM, we consider
the original harmful prompts, rewritten prompts, and rewritten prompts embedded
in nested scenarios. For the adaptive PAIR attack, we use GPT-4o as the attacker
model and run three attack streams for three iterations, allowing at most nine
target-model queries per behavior. We otherwise retain the decoding and
response-level safety-evaluation protocol used in the main experiments.

\begin{table}[t]
\centering
\caption{ASR (\% $\downarrow$) of LLaDA-8B-Instruct under additional jailbreak
attacks.}
\label{tab:additional_jailbreaks}
\setlength{\tabcolsep}{4.0pt}
\renewcommand{\arraystretch}{1.08}
\resizebox{\columnwidth}{!}{%
\begin{tabular}{lcccc}
\toprule
\multirow{2}{*}{Denoising}
& \multicolumn{3}{c}{ReNeLLM}
& \multirow{2}{*}{PAIR} \\
\cmidrule(lr){2-4}
& Original & Rewritten & Nested & \\
\midrule
Vanilla & 0.33 & 0.33 & 0.33 & 48.00 \\
RAEC    & {0.00} & {0.00} & {0.00}
        & {36.00} \\
\bottomrule
\end{tabular}%
}
\end{table}

As shown in Table~\ref{tab:additional_jailbreaks}, the vanilla model already
has a low ASR of 0.33\% on the three ReNeLLM variants, which RAEC further
reduces to zero. More importantly, under the adaptive PAIR attack, RAEC reduces
ASR from 48\% to 36\%, an absolute reduction of 12 percentage points (a 25\%
relative reduction). These results extend the effectiveness of RAEC beyond
DIJA-style attacks while requiring neither retraining nor additional
target-model queries during generation.

\subsection{RAEC Ablation Study}
\label{apen:raec_sensitivity}

We analyze the sensitivity of RAEC to its token set, compliance-token margin
$\delta$, persistence length $L$, early denoising-step window
$\mathcal{T}_{\mathrm{e}}$, early response-position window
$\mathcal{R}_{\mathrm{e}}$, and top-$K$ candidate condition. We report ASR on
DIJA-JBB together with accuracy on AGNews to characterize the safety--utility
trade-off. Unless otherwise specified, the default configuration uses the full
refusal-token set, $\delta=0.001$, $L=3$, the first eight denoising steps, the
first eight response positions, and $K=10$. All remaining RAEC parameters are
held fixed.

For the token-set ablation, \textit{Random} and \textit{Compliance} replace the
refusal-token set with random tokens and compliance tokens, respectively. We
also include a simpler refusal-token-bias baseline using 50\% of the refusal
token set.

\begin{table}[H]
\centering
\caption{Ablation and sensitivity analysis of RAEC on
LLaDA-8B-Instruct. Defaults are shown in bold.}
\label{tab:raec_sensitivity}
\setlength{\tabcolsep}{4pt}
\renewcommand{\arraystretch}{1.04}
\small
\resizebox{\columnwidth}{!}{%
\begin{tabular}{@{}lcc@{}}
\toprule
Setting
& \begin{tabular}[c]{@{}c@{}}DIJA-JBB\\ASR (\% $\downarrow$)\end{tabular}
& \begin{tabular}[c]{@{}c@{}}AGNews\\ACC (\% $\uparrow$)\end{tabular} \\
\midrule
Vanilla (without RAEC) & 8.00 & 82.30 \\

\midrule
\multicolumn{3}{@{}l}{\textit{Token-set choice / baseline}} \\
Random tokens                    & 9.00  & 82.38 \\
Compliance tokens                & 12.00 & 82.80 \\
50\% refusal-token bias          & 7.00  & 82.00 \\
\textbf{Refusal tokens}
& \textbf{6.00} & \textbf{82.20} \\

\midrule
\multicolumn{3}{@{}l}{\textit{Compliance-token margin $\delta$}} \\
$0$                              & 9.00 & 82.40 \\
$\mathbf{0.001}$                 & \textbf{6.00} & \textbf{82.20} \\
$0.002$                          & 6.00 & 82.97 \\
$0.003$                          & 7.00 & 82.30 \\
$0.01$                           & 5.00 & 82.60 \\

\midrule
\multicolumn{3}{@{}l}{\textit{Persistence length $L$}} \\
$1$ (without persistence)        & 7.00 & 82.70 \\
$2$                              & 8.00 & 82.80 \\
$\mathbf{3}$                     & \textbf{6.00} & \textbf{82.20} \\
$4$                              & 5.00 & 82.00 \\

\midrule
\multicolumn{3}{@{}l}{\textit{Early denoising-step window}} \\
$2$                              & 5.00 & 82.90 \\
$4$                              & 5.00 & 81.80 \\
$\mathbf{8}$                     & \textbf{6.00} & \textbf{82.20} \\
$16$                             & 8.00 & 82.90 \\
$64$                             & 8.00 & 82.90 \\

\midrule
\multicolumn{3}{@{}l}{\textit{Early response-position window}} \\
$4$                              & 7.00 & 77.10 \\
$\mathbf{8}$                     & \textbf{6.00} & \textbf{82.20} \\
$16$                             & 7.00 & 80.80 \\
$64$                             & 6.00 & 77.20 \\
$128$                            & 6.00 & 77.20 \\

\midrule
\multicolumn{3}{@{}l}{\textit{Top-$K$ candidate condition}} \\
$5$                              & 7.00 & 82.50 \\
$\mathbf{10}$                    & \textbf{6.00} & \textbf{82.20} \\
$20$                             & 7.00 & 82.80 \\
$50$                             & 9.00 & 82.90 \\
\bottomrule
\end{tabular}%
}
\end{table}

The token-set ablation supports the importance of refusal-specific evidence:
replacing the refusal set with random or compliance tokens increases ASR from
6\% to 9\% and 12\%, respectively, while the simpler 50\% refusal-token-bias
baseline reaches 7\%. Across the remaining hyperparameter sweeps, RAEC achieves
ASRs between 5\% and 9\%, indicating that its safety improvement does not depend
on one isolated parameter value. The response-position window has the clearest
effect on utility: both an overly narrow window and substantially wider windows
reduce AGNews accuracy, whereas the default eight-position window preserves
82.20\% accuracy relative to the vanilla accuracy of 82.30\%.

The step and position defaults are guided by the safety-sensitive region
identified in section~\ref{3.2} and section~\ref{3.3}, rather than being selected
solely to minimize DIJA-JBB ASR. The compliance margin and persistence length
control how strong and stable a refusal signal must be before commitment.
Overall, the default configuration provides a balanced operating point that
improves safety while preserving general-task performance.

\end{document}